\documentclass[11pt]{article}

\usepackage[utf8]{inputenc}
\usepackage[T1]{fontenc}
\usepackage{lmodern}
\DeclareUnicodeCharacter{2010}{-}

\usepackage[a4paper,margin=25mm]{geometry}
\usepackage{amsmath,amssymb}
\usepackage{graphicx}
\usepackage{float}
\usepackage{booktabs}
\usepackage{array}
\usepackage[table]{xcolor}
\usepackage{microtype}
\usepackage[labelfont=bf,font={small,stretch=1.0}]{caption}
\DeclareCaptionLabelSeparator{pipe}{ \textbar\ }
\newfloat{suppfigure}{H}{lsf}
\floatname{suppfigure}{Supplementary Fig.}

\usepackage[numbers,super,sort&compress]{natbib}
\usepackage{setspace}
\usepackage{lineno}
\usepackage[hidelinks]{hyperref}

\graphicspath{{figures/}}

\newcommand*\patchAmsMathEnvironmentForLineno[1]{%
  \expandafter\let\csname old#1\expandafter\endcsname\csname #1\endcsname
  \expandafter\let\csname oldend#1\expandafter\endcsname\csname end#1\endcsname
  \renewenvironment{#1}%
    {\linenomath\csname old#1\endcsname}%
    {\csname oldend#1\endcsname\endlinenomath}}
\newcommand*\patchBothAmsMathEnvironmentsForLineno[1]{%
  \patchAmsMathEnvironmentForLineno{#1}%
  \patchAmsMathEnvironmentForLineno{#1*}}
\AtBeginDocument{%
  \patchBothAmsMathEnvironmentsForLineno{equation}%
  \patchBothAmsMathEnvironmentsForLineno{align}%
  \patchBothAmsMathEnvironmentsForLineno{gather}%
  \patchBothAmsMathEnvironmentsForLineno{multline}%
}

\newcommand{\boxfigof}[4]{%
  \makebox[\textwidth][c]{\setlength{\fboxsep}{6pt}\setlength{\fboxrule}{0.4pt}%
  \fbox{\begin{minipage}{180mm}
    \includegraphics{#2}\par
    \captionof{#1}{#4}\label{#3}
  \end{minipage}}}}
\newcommand{\boxfig}{\boxfigof{figure}}
\newcommand{\boxsfig}{\boxfigof{suppfigure}}

\title{\textbf{Decoupling the clocks of learning and computing in spiking neural networks}}
\author{Michalis Pagkalos \& Panayiota Poirazi}
\date{}
\makeatletter
\renewcommand\@maketitle{%
  \newpage\null\vskip 2em
  \begin{center}%
    {\LARGE \@title \par}%
    \vskip 1.5em
    {\large \@author \par}%
    \vskip 2pt
    {\normalsize Institute of Molecular Biology and Biotechnology,\par}%
    {\normalsize Foundation for Research and Technology-Hellas, Heraklion, Crete, Greece\par}%
    \vskip 6pt
    {\normalsize Correspondence to \href{mailto:poirazi@imbb.forth.gr}{\nolinkurl{poirazi@imbb.forth.gr}}\par}%
  \end{center}%
  \par\vskip 1.5em}
\makeatother

\begin{document}
\maketitle

\begin{abstract}
Biological intelligence thrives on minimal power despite its noisy, imperfect components \cite{niven_energy,faisal_noise,marder_robust}, with neurons that communicate through voltage spikes.
Brain-inspired computing seeks to bring such efficiency to real devices \cite{zador_neuroai,davies_loihi}, which requires training methods tailored to them.
The standard way to train spiking neural networks, backpropagation through time, computes weight updates through every step of a simulated device, so faithful or long simulations can make training costly and unstable \cite{neftci,bengio_longterm,pascanu_difficulty}.
Here we show that spiking networks can learn reliably and efficiently by decoupling the clock of learning from that of computing.
Our method, ChronoProp, calculates weight updates from each neuron's spike counts in fixed time bins using its measured input--output function.
Compared with backpropagation through time, ChronoProp reaches similar accuracy in image classification, with about 3 times less training time and 20 times less memory at the finest timestep tested, and up to 10 times fewer spikes.
On spoken-word benchmarks \cite{cramer_shd}, ChronoProp is far more accurate than backpropagation through time in recurrent networks at fine timesteps, and with simple, hardware-compatible techniques it approaches the best published models \cite{baronig_adaptation}.
These networks also tolerate simulated hardware defects and limitations better than those trained with backpropagation through time.
Therefore, ChronoProp enables stable and efficient training in simulations close to the continuous time of analog hardware.
\end{abstract}

\section*{Introduction}
Evolution has spent hundreds of millions of years shaping biological intelligence to be flexible, robust and efficient \cite{liebeskind_neural_evolution,niven_energy,marder_robust}.
For example, from the moment it hatches, a sea turtle must cross the ocean alone, outswim predators and find food, guided in part by an inherited magnetic map \cite{lohmann_magnetic_maps,gatto_hatchling,mansfield_lost_years}.
It manages all this with a brain weighing less than a gram, in a body that consumes about 20~milliwatts \cite{dodd_loggerhead,wyneken_anatomy,gatto_hatchling}.
By contrast, frontier machine learning research pursues artificial general intelligence mostly by scaling up computation \cite{morris_levels_agi,ho_algoprogress}.
Thus, demand for electricity and chips has soared \cite{iea_energyai,mehonic_masterplan,sevilla_scaling2030}, and the cost of training the largest models has been doubling about every nine months over the past decade \cite{cottier_costs}.
As a result, only a handful of large companies and governments can now afford to train and experiment with such models, leaving academia and most other researchers behind \cite{cottier_costs,ahmed_industry}.

NeuroAI proposes a different path, building intelligent systems inspired by the brain's principles \cite{zador_neuroai,zador_beyond}.
As such, it promises efficiency wherever power, size and response time are limited, for example in prosthetic limbs, mixed-reality headsets, drones and satellites \cite{eshraghian_snntorch,kudithipudi_scale}.
Many neuromorphic chips are designed to deliver that efficiency by running spiking neural networks (SNNs), whose neurons communicate through brief pulses called spikes \cite{davies_loihi,furber_spinnaker,pehle_bss2}.
Such networks are typically trained offline on graphics processing units (GPUs) with backpropagation through time (BPTT) \cite{neftci,muir_commercial}.
BPTT and its variants have greatly expanded what SNNs can learn \cite{eshraghian_snntorch}, but they share limits that stem from how BPTT computes the gradient, the signal that tells each weight how to change.
BPTT differentiates the network's simulation step by step, which ties learning to the simulation's clock.
Its memory and computation therefore grow with the number of steps, hindering efficiency, while multiplying derivatives across so many steps raises the risk of training becoming unstable \cite{bengio_longterm,pascanu_difficulty}.
Yet a simulation approximates a continuous-time system, whether a neuron or a physical device, more closely the finer its timestep is.
BPTT is thus hardest to use exactly where the simulation is most faithful.

Here we introduce ChronoProp (Fig.~\ref{fig:chainrule}), a training method that decouples the clock of learning from that of the simulation.
The network runs at whatever timestep the hardware requires, while the gradient is taken once per time window of many steps.
This idea was inspired by the brain, which computes in continuous time but learns over timescales spanning milliseconds to hours \cite{bi_stdp,gerstner_eligibility,frey_tagging}.
To separate the two clocks, ChronoProp first measures each neuron's frequency--current (FI) curve, the number of spikes it fires in a window as a function of its average input current.
A rectified linear function fitted to this curve is the only model of the device that training needs.
Training then runs the network forward as an ordinary spiking simulation that records only one spike count per neuron per window, and backward as a conventional artificial neural network whose activation function is the fitted curve.

We compare ChronoProp to BPTT in both feedforward and recurrent networks, applied to image and time-series data classification.
On image classification with feedforward networks, ChronoProp reaches accuracy comparable to BPTT's while training faster, with less memory and far less spiking activity, and its savings in training cost grow with the number of steps.
On spoken-word benchmarks, with recurrent networks at a fine timestep of 0.1~ms, ChronoProp is far more accurate than BPTT.
These networks also surpass the most accurate networks that BPTT trains under the same conditions, whatever the connectivity or timestep.
Moreover, ChronoProp readily extends to complex neurons, such as adaptive or dendritic ones, by fitting their spike count with a two-dimensional function.
With a few hardware-compatible techniques, ChronoProp brings simple networks of plain leaky integrate-and-fire (LIF) neurons close to the best published models.
Finally, under simulated hardware defects, networks trained with ChronoProp retain more of their defect-free accuracy than those trained with BPTT, under both hardware-in-the-loop and post-training settings.

\section*{Results}
\subsection*{A gradient per time window, not per step}

To show how the gradient can be taken once over many simulation steps, we compare how BPTT and ChronoProp train the synaptic weights $W$ of an SNN (Fig.~\ref{fig:chainrule}).
Training updates each weight according to the gradient of the loss $\mathcal{L}$, which backpropagation computes by the chain rule, multiplying the derivatives along the chain from the loss back to the weight.

BPTT differentiates the simulation itself, so its chain passes through the membrane potential $U$ and the spike $S$ at every timestep, and the gradient sums over all these steps (Fig.~\ref{fig:chainrule}, right).
The spike is an all-or-none function of $U$, so its derivative $\partial S/\partial U$ is zero everywhere except at the threshold, where it is undefined.
Surrogate-gradient BPTT therefore replaces it with the derivative of a smooth function, the surrogate gradient (Fig.~\ref{fig:chainrule}, right insets) \cite{neftci}.
This method applies to any neuron whose equations are known, but it needs the membrane potential at every step.

ChronoProp forgoes this information and never needs $\partial S/\partial U$, because its chain passes through the spike count $N$ of each neuron in each time window, not through individual spikes (Fig.~\ref{fig:chainrule}, left).
Unlike a spike, the count rises nearly in proportion to the input current $I$ once the neuron starts to fire, so ChronoProp takes $\partial N/\partial I$ from the slope of the neuron's fitted FI curve (Fig.~\ref{fig:chainrule}, left insets).
Training thus needs only this fitted curve and each neuron's count per window, regardless of how many timesteps the window spans.
The membrane potential is simulated at every step but never differentiated.

From ChronoProp's chain we predict four key advantages, each tested in a later section.
First, the backward pass stores one count per spiking neuron instead of the history of every timestep, so training should cost less than with BPTT, by a margin that grows with the number of timesteps (Fig.~\ref{fig:static}).
Second, ChronoProp's chain is shorter than BPTT's and does not lengthen as the timestep shrinks, so recurrent training should remain more stable at a fine timestep (Fig.~\ref{fig:temporal}).
Third, the neuron enters the gradient only through its measured curve, so ChronoProp should extend to more complex neurons by fitting a richer FI curve (Fig.~\ref{fig:adaptive}).
Finally, a defect that merely shifts a spike within its window leaves the count unchanged, so networks trained through such defects should keep more of their accuracy with ChronoProp than with BPTT, whose gradient reads the perturbed membrane potential at every step (Fig.~\ref{fig:defects}).

\begin{figure}[H]
  \boxfig{figures/fig1}{fig:chainrule}{\textbf{How ChronoProp and surrogate-gradient BPTT compute the weight gradient.} Each half shows one neuron, trained with ChronoProp or BPTT, and its boxes are the quantities it computes, from the input spikes $X$, through the input current $I$ (via the weights $W$) and the membrane potential $U$, to its output and the loss $\mathcal{L}$; filled boxes are those the gradient passes through.
  \underline{Left:} ChronoProp learns only from the neuron's spike count $N$ in a time window, so its gradient has three factors, $\partial\mathcal{L}/\partial N \cdot \partial N/\partial I \cdot \partial I/\partial W$, with no sum over time. The middle factor, $f'(I)$, comes from the fitted FI curve and links current to count, so the membrane potential is never differentiated (unfilled box). The upper inset (Fitted proxy) shows counts measured at constant currents (dots) with the fitted curve $N = f(I)$, and the lower inset (Proxy derivative) shows the slope $f'$, which jumps where firing begins.
  \underline{Right:} BPTT differentiates the simulation at every timestep $\Delta t$, through the spike $S$ and the membrane potential, and sums over $t$; here $\partial U/\partial I = R(1-\alpha)$, with $R$ the membrane resistance and $\alpha$ the per-step decay. The upper inset (Spike surrogate) shows the spike as a step at the threshold $\vartheta$ (dark line) and the smooth surrogate $\sigma(U)$ that BPTT substitutes for it (orange), and the lower inset (Surrogate derivative) shows $\sigma'(U)$, which peaks at $\vartheta$.}
\end{figure}

\subsection*{Savings that grow with the step count}

We first tested ChronoProp in its simplest form on image classification, a task with no temporal features, so one time window can span the whole input.
We trained feedforward networks with one hidden layer of LIF neurons on the MNIST \cite{lecun_mnist} and Fashion-MNIST \cite{xiao_fmnist} benchmarks, presenting each image for 50~ms as spikes (Fig.~\ref{fig:static}a).
A finer timestep follows the neuron's dynamics more closely, nearer the continuous time of analog chips \cite{pehle_bss2,richter_dynapse2} (Supplementary Fig.~\ref{sfig:dtlimit}a,b), and lets the rectified linear function fit the FI curve better (Supplementary Fig.~\ref{sfig:dtlimit}c,d).
Yet every added step costs BPTT more, because it differentiates each one, whereas ChronoProp differentiates its hidden layer once per window.

We compared the methods at timesteps $\Delta t$ of 1, 0.5 and 0.1~ms, with the same input and the same non-spiking output layer (see Methods).
Both training methods used the Adam optimizer alone \cite{kingma_adam}, without regularizers or other additions, which would bring further hyperparameters to tune.
The two methods are indistinguishable on MNIST at every timestep, and BPTT is slightly more accurate on Fashion-MNIST, by about 0.4 percentage points (Fig.~\ref{fig:static}b).
Both reach their best validation accuracy at a similar epoch (Supplementary Fig.~\ref{sfig:staticdyn}).

To assess how the training cost grows with the number of steps, we shortened the timestep on MNIST so that the same 50~ms window spans 50--2,000 steps (see Methods).
At 2,000 steps, BPTT takes 2.98 times longer than ChronoProp and uses 21.7 times the memory and 4.17 times the operations (Fig.~\ref{fig:static}c--e).
The ratios for memory and operations keep rising across this range, whereas the ratio for time remains near 3 beyond 500 steps; accuracy stays flat.
With a spiking readout, the gaps in time and memory grow even larger (Supplementary Fig.~\ref{sfig:activity}a,b; see Methods).

Interestingly, ChronoProp's hidden neurons fire at 36--43~Hz at every step count, while BPTT's run at 262--390~Hz, 6 to 10 times more frequently (Fig.~\ref{fig:static}f and Supplementary Fig.~\ref{sfig:activity}c).
With ChronoProp, 29--38\% of hidden neurons fire at least once on a given image, against 47--73\% with BPTT (Supplementary Fig.~\ref{sfig:activity}d).
On event-driven hardware, whose running cost scales with the number of spikes, we expect this substantially sparser activity to save energy both in training and in deployment \cite{davies_loihi}.

\begin{figure}[H]
  \boxfig{figures/fig2}{fig:static}{\textbf{ChronoProp approaches BPTT's accuracy at a fraction of its training cost.} ChronoProp (CP; blue) and BPTT (orange) train the same feedforward spiking network.
\textbf{a} Input encoding: in every 1~ms frame, each pixel spikes with a probability proportional to its intensity.
\textbf{b} Test accuracy on MNIST (squares) and Fashion-MNIST (FMNIST; triangles) at three timesteps $\Delta t$, read at the learning rate and epoch of best validation accuracy. Markers show individual seeds ($n = 10$) and black lines their means. In two-sided paired $t$-tests over seeds, Holm-corrected across the six comparisons, the methods differ on FMNIST (all $P \leq 0.040$) but not on MNIST (all $P \geq 0.75$).
\textbf{c}--\textbf{f} Training cost and hidden-layer activity on MNIST as the number of simulation steps $T$ in the 50~ms window grows from 50 to 2,000 ($\Delta t$ from 1 to 0.025~ms). Panels \textbf{c}--\textbf{f} share the legend in \textbf{c}.
\textbf{c} Time per batch for the forward and backward pass.
\textbf{d} Peak GPU memory per batch over the same passes.
\textbf{e} Effective floating-point operations per sample (EFLOPs: operations whose operands are all non-zero \cite{narduzzi_eflop}), a hardware-independent count.
\textbf{f} Mean hidden-layer firing rate on the test set (logarithmic axis); the grey dashed line marks the shared rate before training.
Green boxes give the BPTT-to-ChronoProp ratio at 500 and 2,000 steps. Error bars show $\pm 1$~s.d. over $n = 3$ networks per method and $T$ (each timed twice in \textbf{c}) and are mostly smaller than the markers; memory did not vary.}
\end{figure}

\subsection*{Stable recurrent training at fine steps}

Time-varying inputs, such as sounds, are a natural fit for SNNs, which integrate signals over time \cite{muir_commercial}.
Yet much of the information in such inputs lies in the timing of their features, which the static version of ChronoProp (Fig.~\ref{fig:static}) ignores.
A single spike count per neuron records how many spikes it fired, but not when.
Temporal ChronoProp captures this timing by counting spikes in successive short windows, or time bins, instead of one long window (Fig.~\ref{fig:temporal}a,b).
These bins can be short enough for the features to stay distinct, but much longer than the simulation step.
Each bin contributes its own gradient, and these add up to one weight update; no gradient crosses from one bin to the next.
Importantly, time is divided only for learning; the network runs as one uninterrupted simulation of the whole input.

To test whether recurrent training stays stable at a fine timestep, as ChronoProp's shorter chain predicts (Fig.~\ref{fig:chainrule}), we trained feedforward and recurrent networks with one hidden layer on the Spiking Heidelberg Digits (SHD) and Spiking Speech Commands (SSC) benchmarks \cite{cramer_shd}, at a coarse (1~ms) and a fine (0.1~ms) timestep.
Before training, we fixed the bin duration $K$ at 10~ms, one membrane time constant and the shortest bin at which the fitted FI curve stays accurate at both timesteps (Supplementary Fig.~\ref{sfig:binning}b,c).
The fine timestep spreads each bin over 100 steps instead of 10 and so resolves its input in more detail (Supplementary Fig.~\ref{sfig:binning}a).
The recurrent networks are narrower, so that every network has the same number of parameters (see Methods).
As with the images, both training methods used Adam alone, each with its default parameters.

Test accuracy depends strongly on the network and the timestep, and each method follows a clear trend (Fig.~\ref{fig:temporal}c--f).
With ChronoProp, accuracy rises with recurrence, and recurrent networks gain further at the fine timestep.
BPTT, by contrast, loses accuracy at the fine timestep, most of all in recurrent networks.
At the coarse timestep, BPTT is more accurate in feedforward networks on both datasets (73.95\% against 67.46\% on SHD and 52.21\% against 48.53\% on SSC) and in recurrent networks on SSC (61.42\% against 56.35\%).
Both the trends and BPTT's wins fit how the two methods compute the gradient.
BPTT is more accurate only where its chain is short or has no loop; around a recurrent loop, it multiplies one derivative per step, and long products of this kind tend to vanish or explode \cite{bengio_longterm,pascanu_difficulty}.
ChronoProp's best condition is thus BPTT's worst.
In recurrent networks at the fine timestep, ChronoProp reaches 80.07\% on SHD and 65.03\% on SSC, against BPTT's 63.60\% and 26.41\% (Fig.~\ref{fig:temporal}d,f).
These are the highest accuracies of the comparison on each dataset, above BPTT's most accurate networks (76.50\% and 61.42\%).

Notably, BPTT's recurrent networks at the fine timestep fall behind because they fail to train, not because they generalize poorly; on SHD they reach only 73.54\% even on the validation set, against ChronoProp's 96.99\% (Fig.~\ref{fig:temporal}c and Supplementary Fig.~\ref{sfig:temporaldyn}b,d).
Over all conditions and learning rates, 41 of BPTT's 400 runs, all recurrent, became unstable, while none of ChronoProp's did.

\begin{figure}[H]
  \boxfig{figures/fig3}{fig:temporal}{\textbf{ChronoProp outperforms BPTT in recurrent networks at a fine timestep.}
\textbf{a} Static ChronoProp takes one gradient over the whole input (upper), and temporal ChronoProp one per time bin (dashed lines; lower).
\textbf{b} One spoken digit from SHD as a raster of input spikes, divided into bins of duration $K$ (dashed lines).
\textbf{c}--\textbf{f} Accuracy on SHD (\textbf{c},\textbf{d}) and SSC (\textbf{e},\textbf{f}), on the validation (left) and test (right) sets. Each panel compares temporal ChronoProp (tCP; blue) and BPTT (orange) in feedforward (FF) and recurrent (REC) networks with the same number of parameters, simulated at a coarse ($\Delta t = 1$~ms) and a fine ($\Delta t = 0.1$~ms) timestep, with bins of $K = 10$~ms. Bars and black lines show means over $n = 10$ seeds, and circles individual seeds. For each method, accuracy is read at the learning rate and epoch of best validation accuracy. The panels share the legend in \textbf{d}.
In two-sided paired $t$-tests over seeds, Holm-corrected across the eight test comparisons, the two methods differ in test accuracy in every condition (all $P \le 7.1 \times 10^{-3}$) except SHD REC at $\Delta t = 1$~ms ($P = 0.24$).}
\end{figure}

\subsection*{One more measurement for richer neurons}

Beyond temporal input, ChronoProp also extends to more complex neurons, by changing how the FI curve is measured and fitted.
Adaptive neurons \cite{bellec_lsnn,yin_adaptive,bittar_garner,baronig_adaptation} need such an extension, because their spike count in a bin depends on more than that bin's input.
Each spike strengthens a long-lasting adaptation current \cite{benda_sfa} that usually decays over many bins, so each bin begins at a different level of adaptation.
A standard FI curve, measured against the input current alone, thus cannot accurately describe these neurons in every bin.

We hypothesized that ChronoProp could train networks of adaptive neurons if each neuron's count were fitted as a function of two variables, its input current and its adaptation current averaged over the bin.
We therefore counted an adaptive neuron's spikes in one bin at many fixed combinations of these two currents, which gives an FI plane (Fig.~\ref{fig:adaptive}a; see Methods).
As in a plain LIF neuron, the count rises with the input current, but a stronger adaptation current shifts the onset of firing to higher currents.
A rectified linear function whose threshold shifts in proportion to the adaptation current describes the spike counts quite accurately ($R^2 = 0.963$ at the fine timestep).
We then trained the networks of Fig.~\ref{fig:temporal} again, unchanged except that every hidden neuron was adaptive.

ChronoProp trains networks of adaptive neurons to a higher test accuracy than plain LIF networks in every condition but SSC at the fine timestep (Fig.~\ref{fig:adaptive}b--e).
As in Fig.~\ref{fig:temporal}, BPTT matches or exceeds ChronoProp wherever its chain is short or has no recurrent loops.
In recurrent networks at the fine timestep, however, ChronoProp reaches 84.03\% on SHD and 63.63\% on SSC, 19.66 and 34.84 percentage points above BPTT.
There, BPTT still fails to fit the training data, whereas ChronoProp again trains stably throughout (Supplementary Fig.~\ref{sfig:adaptivedyn}).

ChronoProp can also train two-compartment neurons, whose dendrite receives independent input and exchanges current with the soma (Supplementary Fig.~\ref{sfig:dendritic}a; see Supplementary Methods).
Their count, measured against the somatic and dendritic currents, again forms an FI plane, which a rectified linear function fits closely ($R^2 = 0.965$ at the fine timestep; Supplementary Fig.~\ref{sfig:dendritic}b).
We trained networks of these neurons to report the sum of two spoken digits from SHD, one at the soma and one at the dendrite (see Supplementary Methods).
The two methods show the same preferences as before.
BPTT is most accurate in feedforward networks at the coarse timestep (43.93\%), and both recurrence and a finer timestep lower its accuracy.
ChronoProp gains from both, and its recurrent networks at the fine timestep are the most accurate of all, reaching 57.69\%, against 19.12\% for BPTT (Supplementary Fig.~\ref{sfig:dendritic}c,d).

One more measured variable thus lets ChronoProp train neurons more complex than plain LIF, and with them it leads BPTT by a wide margin in recurrent networks at the fine timestep.

\begin{figure}[H]
  \boxfig{figures/fig4}{fig:adaptive}{\textbf{ChronoProp extends to neurons with richer dynamics by measuring one more variable.}
\textbf{a} FI plane of an adaptive LIF neuron, whose adaptation current $w$ grows with its spikes and opposes the input. Left, the spike count per bin measured on the simulated neuron, against the input current $I$ and the bin-averaged adaptation current $\bar{w}$. Right, the fitted plane, which ChronoProp differentiates in the backward pass ($g$, gain; $\theta_0$, threshold current without adaptation; $c$, threshold shift per unit of $\bar{w}$).
\textbf{b}--\textbf{e} Accuracy on SHD (\textbf{b},\textbf{c}) and SSC (\textbf{d},\textbf{e}), on the validation (left) and test (right) sets, for temporal ChronoProp (tCP; blue) and BPTT (orange) in the feedforward (FF) and recurrent (REC) networks of Fig.~\ref{fig:temporal}c--f, trained with the same seeds ($n = 10$) and displayed the same way; only the neuron differs. Dashed lines show the mean of the same condition without adaptation (Fig.~\ref{fig:temporal}), so a bar's distance from its dashed line is the effect of adaptation. The panels share the legend in \textbf{c}.
In two-sided paired $t$-tests over seeds, Holm-corrected across the eight test comparisons, the two methods differ in test accuracy in every condition (all $P \leq 0.012$) except SHD FF at $\Delta t = 0.1$~ms ($P = 0.12$) and SHD REC at $\Delta t = 1$~ms ($P = 0.68$).}
\end{figure}

\subsection*{Plain LIF networks rival the best models}

In previous sections, both methods trained networks of fixed width with only the Adam optimizer, so that the comparisons concerned the learning rules alone.
Accuracies therefore stayed well below the best published ones, which leaves open whether a gradient taken from spike counts alone limits network performance.
ChronoProp's recurrent networks at the fine timestep reached 96.99\% on the SHD validation set but only 80.07\% on the test set (Fig.~\ref{fig:temporal}c,d), a sign that generalization, not learning, limited them.
On SSC, a harder task of 35 spoken commands, they had not yet fitted even the training data, which suggested that this task needs larger networks.

To improve generalization, we added three simple, hardware-compatible techniques (Supplementary Fig.~\ref{sfig:tricks}).
\emph{Receptive fields} restrict each neuron to a band of neighbouring input channels, much as auditory neurons respond to a narrow range of frequencies.
In each training sample, \emph{dropout} silences a random set of neurons and \emph{channel jitter} shifts the input by a random number of channels.
We then varied the hidden layer's width and recurrent connection density, to find how accurate one layer of plain LIF neurons, with only its weights trained, can become.

All networks ran at the fine timestep, where ChronoProp works best (Fig.~\ref{fig:temporal}).
As a reference, we used the most accurate published SNNs, which were trained with BPTT and have more layers, richer neuron dynamics or learned neuron parameters in addition to weights.
These models also differ from ours in timestep, training length and data augmentation, so the comparison shows how close one plain layer comes rather than which method is more accurate.
Following Baronig et al.~\cite{baronig_adaptation}, we included only SHD results that were validated, like ours, on data held out from the training set.

Within the ranges we tested, recurrence matters far more than width (Fig.~\ref{fig:recipe}), most likely because no gradient crosses a ChronoProp bin, so information that spans bins must be held in recurrent activity.
On SHD, the best network reaches 92.81\% with 96,256 parameters, and a smaller one, with a quarter of the parameters (24,064), still reaches 91.53\%.
Both are close to the published models validated the same way, the most accurate of which reach 93.18\% with one layer of 37,500 parameters and 93.79\% with two layers of 0.45 million (Fig.~\ref{fig:recipe}a).
On SSC, the best network is also the widest we trained (2,048 neurons) and reaches 74.25\%, level with one published model (74.2\%) and below the others, which reach up to 80.7\% (Fig.~\ref{fig:recipe}b).

Even with a gradient taken from spike counts alone, one hidden layer of plain LIF neurons thus approaches models with more layers, richer neuron dynamics or learned neuron parameters.

\begin{figure}[H]
  \boxfig{figures/fig5}{fig:recipe}{\textbf{With a few hardware-compatible techniques, ChronoProp trains networks with one hidden layer of LIF neurons to competitive accuracy.}
\textbf{a},\textbf{b} Test accuracy on SHD (\textbf{a}) and SSC (\textbf{b}) against the number of trainable parameters (log scale). The networks are simulated at $\Delta t = 0.1$~ms and trained with ChronoProp and the techniques of Supplementary Fig.~\ref{sfig:tricks}. Each line is one width $N$ of the hidden layer, marked with the same shape and tint of blue in both panels, darkest for the widest. Its points are recurrent connection densities (the fraction of possible recurrent connections present), printed beside each marker; parameters grow with density. In \textbf{b}, the widest networks ($N = 2048$) are trained only at densities 0, 0.25 and 0.5. Markers and error bars show mean $\pm$ s.d. over $n = 10$ seeds, read at the epoch of best validation accuracy. Orange diamonds are published spiking models, labelled as in the comparison table of Baronig et al.~\cite{baronig_adaptation}, with error bars where that table gives them; their neuron models, depth, training length and augmentation differ from ours. BHRF, balanced harmonic resonate-and-fire neurons; SE-adLIF and EF-adLIF, adaptive LIF neurons discretized with the symplectic and the forward Euler method; LIF, plain LIF neurons; ALIF, LIF neurons with threshold adaptation; RadLIF, adaptive LIF neurons with batch normalization; cAdLIF, constrained adaptive LIF neurons; DCLS-Delays, feedforward networks with learned synaptic delays. The dotted lines at 90\% (\textbf{a}) and 70\% (\textbf{b}) serve only as visual guides. The panels share the legend in \textbf{b}.}
\end{figure}

\subsection*{Robust training under simulated defects}

The networks so far were trained on defect-free simulations; but on a chip, membrane potentials are noisy, weights are programmed imprecisely and stored in few bits, and membrane time constants vary across neurons ($\tau$ mismatch).
We therefore studied how ChronoProp trains under each of these four defects, one at a time (Supplementary Fig.~\ref{sfig:defectmech}).
Because ChronoProp's gradient reads spike counts rather than the membrane potential of every neuron at every step, we expected networks trained with it to keep more of their accuracy under these defects than with BPTT, even with an FI curve measured once, before training (see Methods).

To test this, we trained the best SHD network of Fig.~\ref{fig:recipe}a, with its three techniques, in two settings (see Methods).
Simulated hardware-in-the-loop training mimics having a chip available during training, so the network is trained and tested with the defect present.
Post-training perturbation mimics transferring a network trained without defects to a chip, so it meets the defect only at test.
We compared ChronoProp with two controls.
The first is BPTT on the same recurrent network at the fine timestep, where BPTT trained worst (Fig.~\ref{fig:temporal}).
At a learning rate chosen for stability (see Methods), BPTT reaches 77.16\% without defects, against ChronoProp's 92.81\%.
The second control, BPTT*, removes the recurrent connections and trains at the coarse timestep, where BPTT is most stable, and reaches 83.40\%.
Because the methods start from different accuracies, we compared retained accuracy, namely the percentage of defect-free accuracy kept under a defect.

With hardware-in-the-loop training, every method is robust to $\tau$ mismatch up to a coefficient of variation of 30\% (Fig.~\ref{fig:defects}c).
This is most likely because the mismatch lies in the leak conductance, which sets both $\tau$ and the membrane resistance, so each input still moves the membrane potential by the same amount.
The other three defects separate the methods, and ChronoProp retains more of its accuracy than BPTT at every level (Fig.~\ref{fig:defects}a,b,d).
Under membrane noise as strong as three firing thresholds, ChronoProp networks keep 95.9\% of their accuracy, against 57.1\% with BPTT.
With two-bit weights, they keep 93.5\%, against 73.1\%.
BPTT* loses less, yet even against it ChronoProp retains more wherever a defect costs BPTT* at least 5\% of its accuracy, except under the strongest weight noise.

With post-training perturbation, the methods generally retain less than with hardware-in-the-loop training (Supplementary Fig.~\ref{sfig:defectptp}), but ChronoProp's networks still retain more than BPTT's at 27 of 29 defect levels.
Against BPTT*, they retain more at every level of weight noise and at seven bits or fewer, and less only under strong membrane noise.
With four-bit weights, they keep 91.5\%, against 36.4\% with BPTT and 46.8\% with BPTT*.

ChronoProp thus trains through simulated hardware defects from spike counts and an FI curve measured once, and its networks keep far more of their accuracy than BPTT's, whether they meet the defect in training or only at test.

\begin{figure}[H]
  \boxfig{figures/fig6}{fig:defects}{\textbf{ChronoProp trains through simulated hardware defects and retains more of its defect-free accuracy than BPTT.}
Networks are trained on SHD with the defect present and tested at the same level (hardware-in-the-loop training). Each curve shows retained accuracy, the test accuracy as a percentage of the defect-free accuracy of the same method and initialization (dotted grey line, 100\%). Temporal ChronoProp (tCP; blue) and BPTT (orange) train the best network of Fig.~\ref{fig:recipe}a ($N = 512$, recurrent density 0.25) at $\Delta t = 0.1$~ms. BPTT* (red) trains it without its recurrent connections at $\Delta t = 1$~ms, where BPTT is most stable (Fig.~\ref{fig:temporal}).
\textbf{a} Membrane noise, in multiples of the firing threshold.
\textbf{b} Weight noise, the programming error of each weight (\% of full scale).
\textbf{c} $\tau$ mismatch, the coefficient of variation (CV) of the membrane time constant $\tau$ across neurons. The spread arises in the leak conductance, so the membrane resistance scales with $\tau$ at fixed capacitance.
\textbf{d} Weight quantization, with fewer bits towards the right.
Points show means and bands $\pm 1$~s.d. over networks and simulated chips ($n = 10$ networks per point, each scored on 20 unseen chips in \textbf{b}, with 5 noise realizations in \textbf{a}, and once in \textbf{c},\textbf{d}). ChronoProp retains more than BPTT at every level in \textbf{a}, \textbf{b} and \textbf{d}, and more than BPTT* wherever BPTT* retains 95\% or less, except at 30\% weight noise (two-sided paired $t$-tests on per-network means, Holm-corrected across the 22 levels for BPTT and all 29 for BPTT*; all $P < 0.05$). The panels share the legend in \textbf{d}.}
\end{figure}

\section*{Discussion}
Our results show that SNNs can learn stably and efficiently at a slower pace than they compute, much as the brain learns from noisy, imperfect neurons and synapses over timescales of milliseconds to hours \cite{faisal_noise,marder_robust,bi_stdp,gerstner_eligibility,frey_tagging}.
Despite rapid progress in SNN training, learning usually remains tied to a step-by-step simulation of the device, with its neuron equations and idealized parts.
ChronoProp loosens these ties by taking one gradient per time bin from the device's own spike counts and an FI curve measured on its neurons.

Training takes less time and memory, by margins that grow with every simulated step, so more experiments fit into the same GPU hours.
Networks also fire sparsely without a regularizer, which should save energy on event-driven chips (Fig.~\ref{fig:static}).
Recurrent networks can be trained stably on long sequences such as speech, and reach their best accuracy at the fine timesteps where BPTT often fails (Fig.~\ref{fig:temporal}).
Some richer neurons, such as adaptive and dendritic ones, can be trained by measuring their curve along each added variable (Fig.~\ref{fig:adaptive} and Supplementary Fig.~\ref{sfig:dendritic}).
With plain LIF neurons, recurrent connections and a few hardware-compatible techniques, simple networks come close to the most accurate published models (Fig.~\ref{fig:recipe}).
Because these networks train only their weights, with no adaptation, trained time constants or delays, they should also be simpler to build on a chip.
Finally, under simulated defects, networks keep more of their accuracy with ChronoProp than with BPTT, which suggests that a network could be trained for the chip it will run on (Fig.~\ref{fig:defects}).

ChronoProp is not meant to replace BPTT but to train recurrent networks that run over many steps, chips that report only spikes, and neurons whose equations are hard to differentiate.
BPTT is the better choice where its chain of derivatives is short or has no loop, as in feedforward networks, especially at coarse timesteps (Figs.~\ref{fig:static} and~\ref{fig:temporal}).
To compare the learning rules alone, we trained both methods with the Adam optimizer and nothing else.
While gradient clipping often stabilizes training \cite{pascanu_difficulty}, adding it to BPTT left these networks near chance accuracy in preliminary runs.
Other additions, such as an activity regularizer, may recover part of BPTT's accuracy, but each brings further settings to tune.
On chips that record the membrane potential throughout a run \cite{cramer_pnas}, BPTT gains richer information, but its cost grows with the run's length.
ChronoProp's own price is that spike timing within a bin does not enter the gradient, and that each variable added to the curve must be recorded during training, as the adaptation current was here.
Moreover, we did not test deeper networks, although the backward pass extends to more layers.
All our results come from simulation, and we have yet to test them on a physical chip.

Other methods also train SNNs without BPTT's step-by-step gradient, and Extended Data Table~1 compares them with ChronoProp.
The most similar methods pair spiking activity with an artificial network, either in the loop on a wafer-scale neuromorphic system \cite{schmitt_hil} or layer by layer in simulation \cite{wu_tandem}.
Both train feedforward networks on static or short input through an activation that is assumed or derived.
Spike-train backpropagation trains recurrent networks through a derived relation that reads spike times \cite{zhang_strsbp}.
Rate-based backpropagation reduces BPTT's cost by approximating its gradient \cite{yu_ratebp}.
The Neural Engineering Framework can measure neurons on a chip but sets its weights with linear decoders \cite{eliasmith_nef,braindrop}.
ChronoProp differs in two ways.
It trains recurrent networks from spike counts, on time-varying input over many steps, and it updates their hidden and recurrent weights through an FI curve measured on the neuron rather than derived from its equations.

ChronoProp can be adapted to the task and to the neuron through two choices, how spikes are counted and which variables the measured curve depends on.
Here, spikes were counted over the whole input or per time bin, and the curve was fitted along one or two variables.
This suggests three research directions.
First, a network could use time bins of different durations to learn over multiple timescales, from fine detail to whole words, as biological intelligence does.
Second, since no gradient crosses a bin, computing a loss at every bin would allow online learning, with training memory and the cost of each backward pass independent of the input's length.
Finally, measuring further variables or fitting other functions could extend ChronoProp to neurons with more compartments and active dendrites.
In principle, it applies to any device whose spikes can be counted and whose connections can be set.

Whether NeuroAI delivers the efficiency it promises will depend on more than benchmark scores.
It will also depend on what training costs and how much it demands of the hardware \cite{momeni_pnn}.
ChronoProp needs little more than a measured curve and a count of spikes.
That already saves time and memory on today's GPUs, and we hope it will bring new neuron models and new chips within reach of the tools of deep learning.
Perhaps machines, like brains, do not need to learn as fast as they compute.

\section*{Methods}
\subsection*{Neuron model and simulation}

Every spiking neuron in this study, apart from the adaptive neurons of Fig.~\ref{fig:adaptive} and the two-compartment neurons of Supplementary Fig.~\ref{sfig:dendritic}, is a current-based leaky integrate-and-fire (LIF) neuron~\cite{neftci}, simulated with the exponential Euler method at a timestep $\Delta t$.
At step $t$, neuron $i$ of a hidden layer updates its synaptic current $I_i$, its membrane potential $U_i$ and its spike $S_i$ as
\begin{align}
I_i[t] &= \beta\, I_i[t-1] + \sum_j W_{ij}\, X_j[t] + \sum_k W^\mathrm{rec}_{ik}\, S_k[t-1], \label{eq:syn}\\
\hat U_i[t] &= \alpha\, U_i[t-1] + R\,(1-\alpha)\, I_i[t], \label{eq:mem}\\
S_i[t] &= \Theta\bigl(\hat U_i[t]-\vartheta\bigr), \qquad U_i[t] = \begin{cases} 0 & \text{if } S_i[t] = 1,\\ \hat U_i[t] & \text{otherwise,} \end{cases} \label{eq:spike}
\end{align}
where $\hat U_i$ is the membrane potential before the reset, $X_j[t]$ the number of input spikes from input $j$ at step $t$, $W$ the input weights, $W^\mathrm{rec}$ the recurrent weights (absent in feedforward networks), $\Theta$ the Heaviside step function with $\Theta(0)=0$, and $\alpha = e^{-\Delta t/\tau_\mathrm{mem}}$ and $\beta = e^{-\Delta t/\tau_\mathrm{syn}}$ the per-step decays of the membrane potential and the synaptic current, respectively.
All neurons share the same parameters ($\tau_\mathrm{mem} = 10$~ms, $\tau_\mathrm{syn} = 2$~ms, membrane resistance $R = 5$, threshold $\vartheta = 1$).
The two time constants are of the order of those of biological neurons and synapses; we did not tune any of these values.
Each neuron starts every sample at rest ($U = I = 0$), resets to zero after a spike and has no refractory period.

The output layer has one non-spiking neuron $m$ per class, which receives the hidden layer's spikes through the output weights $W^\mathrm{out}$; it is the same neuron with its threshold raised to $10^9$, so it never fires, and its membrane potential integrates its input with leak but without reset.
Each class's logit is the membrane potential of its output neuron averaged over all steps of the sample, and the loss is the cross-entropy of the softmax over the logits.

Weights are drawn uniformly between $-1/\sqrt{n_\mathrm{in}}$ and $1/\sqrt{n_\mathrm{in}}$, where $n_\mathrm{in}$ is the number of inputs a neuron receives, and no layer has biases.
With the same seed, the networks trained with ChronoProp and with BPTT start from identical weights.
We implemented all networks in PyTorch~\cite{paszke_pytorch} and compiled them with TorchScript.

\subsection*{ChronoProp}

To measure the FI curve that enters ChronoProp's gradient, we drove a single neuron from rest with a constant current $I$ for the duration of one bin, the time window over which ChronoProp counts spikes ($K/\Delta t$ steps at the training timestep) and counted its spikes.
The current is added to $I_i[t]$ in equation~\eqref{eq:mem} at every step, without passing through the synaptic decay of equation~\eqref{eq:syn}.
We repeated this at 201 currents evenly spaced from 0 to 1 and fitted the counts by least squares with the rectified linear function
\begin{equation}
f(I) = \max\bigl(0,\; g\,(I-\theta_0)\bigr),
\label{eq:fi}
\end{equation}
whose two parameters are the gain $g$ and the threshold current $\theta_0$.
The simulated neuron is deterministic, so one measurement before training suffices for each timestep and bin duration.

ChronoProp divides each sample into bins of duration $K$.
A short bin keeps brief temporal features in separate bins, but the bin must be long enough for the spike count to follow the input current.
We set $K = 10$~ms, one membrane time constant, and held it fixed, without tuning, in every network trained on audio.
It is the shortest bin at which the fitted FI curve reaches $R^2$ above 0.94 at both timesteps (0.9464 at the coarse and 0.9655 at the fine timestep; Supplementary Fig.~\ref{sfig:binning}c).
In its static form (Fig.~\ref{fig:static}), ChronoProp uses a single bin that spans the whole 50~ms input.

In the forward pass, ChronoProp simulates the network step by step, without computing gradients and without interruption at the bin boundaries, and records the spike count $N_i[b]$ of every hidden neuron in every bin $b$.
The backward pass treats each hidden layer as a conventional layer with a rectified linear activation, whose input is
\begin{equation}
a_i[b] = \sum_j W_{ij}\, N^\mathrm{in}_j[b] + \sum_k W^\mathrm{rec}_{ik}\, N_k[b],
\label{eq:proxy_in}
\end{equation}
where $N^\mathrm{in}_j[b]$ is the number of input spikes from input $j$ in bin $b$.
Its output, however, is the measured count $N_i[b]$, not the value of the fitted function, so the loss is computed on the spiking network's own activity.
The recurrent weights $W^\mathrm{rec}$ receive their gradient through equation~\eqref{eq:proxy_in}, but the counts $N_k[b]$ they multiply enter as constants, so the backward pass contains no loop through time.

The derivative of this activation is the slope of the fitted curve, read at the constant current that matches the neuron's input.
The curve was measured with a constant current $I$, which adds the same input at each of the bin's $K/\Delta t$ steps, $IK/\Delta t$ in total.
A spike arriving through a weight $W_{ij}$ instead enters the synaptic current once, and its contribution then decays by a factor $\beta$ at every step, so over the bin it adds only about $W_{ij}(1+\beta+\beta^2+\dots) = W_{ij}/(1-\beta)$, since $0 < \beta < 1$.
We therefore read the curve at $a_i[b]/\kappa$, the constant current that would deliver the same total input over the bin, with $\kappa = (K/\Delta t)(1-\beta)$.
Read at $a_i[b]$ itself, the curve would treat every spike as a current held for the whole bin, overestimate each neuron's drive about $\kappa$-fold and pass gradient to neurons whose input never reached threshold.
With the conversion, the derivative is
\begin{equation}
\frac{\partial N_i[b]}{\partial a_i[b]} = \frac{g}{\kappa}\,\Theta\bigl(a_i[b]-\kappa\,\theta_0\bigr).
\label{eq:proxy_grad}
\end{equation}

Once the hidden layer has completed bin $b$, the output neurons are simulated over the same bin, driven by the constant current $\sum_i W^\mathrm{out}_{mi}\, N_i[b]$, added like the measurement current, and their membrane potential is differentiated at every step of the bin.
Each bin starts from the output neurons' membrane potential at the end of the previous bin, taken as a constant.
Gradients thus never cross a bin boundary, and the gradient of the loss is a sum of one term per bin, applied as one weight update per batch.

\subsection*{BPTT baseline}

As the baseline, we trained the same networks with surrogate-gradient BPTT~\cite{neftci}, which differentiates the whole simulation of equations~\eqref{eq:syn}--\eqref{eq:spike}.
In the backward pass, the derivative of the spike, $\partial S/\partial \hat U$, is replaced by the derivative of a scaled arctangent~\cite{fang_plif},
\begin{equation}
\sigma'(\hat U) = \frac{\gamma/2}{1 + \bigl(\pi \gamma\,(\hat U-\vartheta)/2\bigr)^2},
\label{eq:surrogate}
\end{equation}
with steepness $\gamma = 2$ in all networks trained on audio and $\gamma = 8$ in Fig.~\ref{fig:static}, each the steepness that gave the best accuracy in preliminary runs.
No gradient passes through the reset.
Neither method used gradient clipping, and neither used a regularizer except in Figs.~\ref{fig:recipe} and \ref{fig:defects}, whose networks train with dropout and channel jitter.

\subsection*{Datasets and input encoding}

For Fig.~\ref{fig:static}, we scaled the pixel intensities $x_j$ of MNIST~\cite{lecun_mnist} and Fashion-MNIST~\cite{xiao_fmnist} to the range 0--1 and presented each image for 50~ms as input spikes.
In every 1~ms frame, pixel $j$ spikes with probability $0.2\,x_j$, a rate of 200~Hz at full intensity, and the spike enters the network on the opening step of the frame, so the input is the same whatever the timestep.
We held out 20\% of each training set, stratified by label, for validation (48{,}000 training and 12{,}000 validation images) and tested on the 10{,}000 test images.
The networks of Fig.~\ref{fig:static} have 784 inputs, one feedforward hidden layer of 128 neurons and 10 output neurons.
For the accuracy comparison (Fig.~\ref{fig:static}b and Supplementary Fig.~\ref{sfig:staticdyn}), both methods tried seven learning rates from $10^{-5}$ to $10^{-2}$, spaced by a factor of about three.

SHD and SSC~\cite{cramer_shd} contain recordings of spoken digits (20 classes) and spoken commands (35 classes), converted into spikes on 700 input channels.
We counted each channel's spikes in consecutive audio frames and presented one frame per simulation step, 10~ms frames at the coarse timestep ($\Delta t = 1$~ms) and 1~ms frames at the fine one ($\Delta t = 0.1$~ms).
The recordings thus play ten times faster than real time at both timesteps, and each 10~ms bin covers the same 100~ms of audio (Supplementary Fig.~\ref{sfig:binning}a).
Each batch is padded with zeros to its longest recording, and for ChronoProp further to a whole number of bins.
SSC provides its own validation set (75{,}466 training, 9{,}981 validation and 20{,}382 test recordings).
SHD does not, so we held out 15\% of its training set, stratified by label (6{,}932 training, 1{,}224 validation and 2{,}264 test recordings).

\subsection*{Training and evaluation}

Both methods trained with the Adam optimizer~\cite{kingma_adam} at PyTorch's default settings and without weight decay, on batches of 256 images or 128 recordings.
In Figs.~\ref{fig:static}--\ref{fig:adaptive} and Supplementary Fig.~\ref{sfig:dendritic}, the learning rate stayed constant.
In Figs.~\ref{fig:recipe} and \ref{fig:defects}, the learning rate increased linearly over the first 10 epochs, from a thousandth of its peak value to the peak, and then decreased along a cosine to a tenth of the peak at the last epoch.
Every run trained for a fixed number of epochs without early stopping, 120 for the accuracy comparison of Fig.~\ref{fig:static}, 400 on SHD and 150 on SSC in Figs.~\ref{fig:temporal}--\ref{fig:defects}, and 500 in Supplementary Fig.~\ref{sfig:dendritic}.
We measured validation and test accuracy after every epoch and report test accuracy at the epoch of best validation accuracy.
For each method and condition, we selected the learning rate whose runs reached the best validation accuracy on average over seeds, from the rates given with each figure's networks.
We fixed several other settings in preliminary runs, which are not among the reported results; each is named where it applies.

\subsection*{Training-cost benchmark}

To measure how the training cost grows with the number of steps $T$, we trained further networks on MNIST at seven timesteps from 1 to 0.025~ms ($T = 50$, 100, 200, 500, 1{,}000, 1{,}500 and 2{,}000), keeping the input frames at 1~ms.
Each method trained for 60 epochs with three seeds at three learning rates ($10^{-4}$, $3\times10^{-4}$ and $10^{-3}$ for ChronoProp; $3\times10^{-4}$, $10^{-3}$ and $3\times10^{-3}$ for BPTT), and for each $T$ we kept the rate with the highest mean validation accuracy.
BPTT selected its largest rate at four step counts; $10^{-2}$ gave lower validation accuracy wherever we tested it.
We measured each network after its last epoch, one at a time on a single NVIDIA RTX 2080 Ti GPU, compiled with TorchScript as in training.
Training time is the mean duration of a batch of 256 images through the forward pass, the loss and the backward pass, over 187 batches after 10 warm-up batches, timed twice per network, and peak memory is the largest amount of GPU memory allocated in any batch.

To count operations independently of the hardware, we used the effective floating-point operations (EFLOPs) of Narduzzi et al.~\cite{narduzzi_eflop}, which count only operations whose operands are all non-zero.
Their rules cover inference only, and we extended them to the backward pass; like theirs, our count leaves out the cost of moving data.
We replayed each trained network's forward and backward passes on its actual tensors for the 10{,}000 MNIST test images and report the convention that charges each added term, which also gives the smallest BPTT-to-ChronoProp ratio of the five we computed (Supplementary Methods).

The firing rates of Fig.~\ref{fig:static}f are the hidden spikes per test image counted in this replay, divided by the 128 hidden neurons and the 50~ms window.
We also recorded them before training and every 5 epochs during it (Supplementary Fig.~\ref{sfig:activity}c); they had not yet settled after 60 epochs.
For Supplementary Fig.~\ref{sfig:activity}a,b, we copied the trained weights into networks whose output neurons spike and measured only their time and memory; we did not train such networks, because ChronoProp's gradient through spiking output neurons is zero at initialization, where no output neuron's input reaches the threshold of its fitted curve.

\subsection*{Audio networks and training stability}

In Fig.~\ref{fig:temporal}, the feedforward networks have one hidden layer of 256 neurons.
The recurrent hidden layer connects every neuron to every other and to itself, and it is narrower, so that the input, recurrent and output weights add up to nearly the same number of parameters, with 200 neurons on SHD (184{,}000 parameters against 184{,}320) and 201 on SSC (188{,}136 against 188{,}160).
Each method tried five learning rates spaced by a factor of about three, from $1.5\times10^{-5}$ to $1.5\times10^{-3}$ for ChronoProp and from $3\times10^{-5}$ to $3\times10^{-3}$ for BPTT, the same in all eight conditions.

We call a run unstable if its training loss, averaged over the batches of each epoch, ends training more than 0.5 above its lowest value, and at chance if it ends above $0.95\ln C$, where $\ln C$ is the loss of a uniform prediction over $C$ classes.
None of ChronoProp's 100 runs at rates at or above BPTT's selected rate became unstable, whereas 7 of BPTT's 160 runs at rates at or below ChronoProp's selected rate did.
Restricting BPTT to ChronoProp's selected rate would lower its validation accuracy in recurrent networks by 6.64--22.61 percentage points.

\subsection*{Adaptive neurons}

The adaptive neurons of Fig.~\ref{fig:adaptive} carry an adaptation current $w_i$, which enters equation~\eqref{eq:mem} as an extra term, $-(1-\alpha)\,w_i[t-1]$, rises by $B$ with each spike and otherwise decays with the time constant $\tau_w$,
\begin{equation}
w_i[t] = e^{-\Delta t/\tau_w}\, w_i[t-1] + B\, S_i[t].
\label{eq:adapt}
\end{equation}
We set $B = 2$ and $\tau_w = 150$~ms in preliminary ChronoProp runs on SHD and did not tune them further.

To measure the FI plane, we held $w$ fixed throughout one bin, started the neuron from rest, counted its spikes at 201 input currents from 0 to 1 and 101 values of $w$ from 0 to 5, and fitted the counts by bounded least squares with
\begin{equation}
f(I, \bar w) = \max\bigl(0,\; g\,(I-\theta_0-c\,\bar w)\bigr),
\label{eq:fi_plane}
\end{equation}
which at the fine timestep gives $g = 5.268$, $\theta_0 = 0.230$ and $c = 0.2000$, matching $1/R$ ($R^2 = 0.963$).
In training, ChronoProp's forward pass records each hidden neuron's adaptation current averaged over the bin, $\bar w_i[b]$, which places the neuron on the plane, and the derivative becomes
\begin{equation}
\frac{\partial N_i[b]}{\partial a_i[b]} = \frac{g}{\kappa}\,\Theta\bigl(a_i[b]-\kappa\,\theta_0-\kappa\,c\,\bar w_i[b]\bigr),
\label{eq:proxy_adapt}
\end{equation}
with $\bar w_i[b]$ entering as a constant.
The conversion by $\kappa$ matters here, because the adaptation term differs between neurons and bins; without it, adaptation would shift each threshold about $\kappa$ times too little.
BPTT differentiates through $w$ as through the other state variables.
The networks, seeds and training settings are those of Fig.~\ref{fig:temporal}, except that ChronoProp tried four learning rates from $5\times10^{-5}$ to $1.5\times10^{-3}$.
The two-compartment neurons and the addSHD task of Supplementary Fig.~\ref{sfig:dendritic} are described in Supplementary Methods.

\subsection*{Receptive fields, dropout and channel jitter}

The networks of Fig.~\ref{fig:recipe} run at the fine timestep with one hidden layer of LIF neurons, the non-spiking output layer and the techniques below.

\textbf{Receptive fields.} Before training, we summed the 700 input channels in groups of five neighbours, which leaves 140 channels.
Each hidden neuron receives only 40 adjacent channels, and the windows of the $N$ neurons are spread evenly across the range (neuron $i$ starts at channel $\mathrm{round}\bigl(i\,(140-40)/(N-1)\bigr)$).
These weights start uniform between $\pm 1/\sqrt{40\cdot 5}$, and all other input weights stay zero throughout training.

\textbf{Sparse recurrence.} Each recurrent connection is present with probability $\rho$, the connection density, so a neuron receives about $\rho N$ of them, with weights starting uniform between $\pm 1/\sqrt{\rho N}$.
The pattern is drawn once per network.

\textbf{Dropout.} Each hidden neuron is silenced with probability 0.025 for a whole training sample.
A silenced neuron keeps integrating its input, but its spikes do not reach the output layer, the recurrent connections or the bin count, and the other neurons' spikes are not rescaled.
The weights into it still receive a gradient, because the backward pass computes its input from the incoming counts (equation~\eqref{eq:proxy_in}); the weights leaving it receive none.

\textbf{Channel jitter.} Each training sample is shifted as a whole along the channel axis by $\mathrm{round}(4z)$ channels, with $z$ drawn from a standard normal distribution for every sample and epoch, and the vacated channels are filled with zeros.
Dropout and jitter act only in training.

\textbf{Network sizes.} On SHD, we trained hidden layers of 128, 256 and 512 neurons at densities $\rho$ of 0, 0.25, 0.5, 0.75 and 1.
On SSC, where the networks of Fig.~\ref{fig:temporal} had not yet fitted the training data, we trained wider ones, of 256, 512 and 1{,}024 neurons at the same densities and 2{,}048 at densities up to 0.5.
The parameter count, $40N + \rho N^2 + CN$ for $C$ classes, takes the recurrent connections at their expected number.
The published models we compare with are those of Table~2 of Baronig et al.~\cite{baronig_adaptation}.

\textbf{Settings and learning rates.} The technique settings and the learning rates come from preliminary runs that compared test accuracy.
We set the jitter and the dropout probability on SHD, in networks of 256 neurons at density 1.
The grouping of five channels and the receptive-field width had to give the best test accuracy on SSC as well, and the three rates, $5\times10^{-5}$, $10^{-4}$ and $3\times10^{-4}$, were the best set on six SHD and six SSC conditions.
Each condition then selected its rate from this set on validation.
The largest rate was selected in every SHD condition.
On SSC, $10^{-4}$ or $3\times10^{-4}$ was selected in every condition except 2{,}048 neurons at density 0.5, where the smallest rate was selected.

\subsection*{Hardware defects}

\textbf{Networks.} With ChronoProp, we trained the best SHD network of Fig.~\ref{fig:recipe} (512 neurons at density 0.25) with the settings and the learning rate of Fig.~\ref{fig:recipe} ($3\times10^{-4}$).
We trained the same network with BPTT, from the same initial weights and connection patterns.
At its validation-selected rate in preliminary runs ($10^{-2}$; three seeds), its training loss repeatedly rose sharply, so we set the rate one decade lower, where it no longer did, at a cost of about one percentage point of test accuracy (78.22\% against 79.23\%).
BPTT* trained the network without its recurrent connections (30{,}720 parameters) at the coarse timestep, at $10^{-3}$, the rate it selected on validation in preliminary runs; at the same seed, its input weights start from a different draw than in the other two methods.

\textbf{Membrane noise.} We added an independent Gaussian term to the membrane potential of every hidden and output neuron at every step of every sample, after equation~\eqref{eq:mem} and before the threshold.
Its level is the s.d. $\sigma_V$ that the noise alone would produce in a membrane without input or reset, in multiples of the threshold; the s.d. per step, $\sigma_V\sqrt{1-\alpha^2}$, keeps this level the same at both timesteps.

\textbf{Weight noise.} To emulate programming error~\cite{ambrogio_analog,joksas_nonideality}, we clipped each weight $W_{ij}$ to a full-scale range $\pm F$ and then perturbed it by Gaussian noise of s.d. $pF$, where $p$ is the level.
We chose $F$ for each array of $n$ weights to minimize the expected squared error $\sum (|W_{ij}| - F)_+^2 + n\,p^2F^2$, which balances clipping against noise and has an exact solution.

\textbf{Quantization.} We rounded each weight to one of $2^{n_\mathrm{bit}}-1$ evenly spaced levels between $-F$ and $F$, with $F$ chosen for each array to minimize the squared rounding error~\cite{nagel_quantization}.
Weight noise and quantization act separately on the input, recurrent and output weights, and only on connections that exist.

\textbf{$\tau$ mismatch.} To emulate device mismatch~\cite{buchel_mismatch}, each neuron's $\tau_\mathrm{mem}$ and $R$ are multiplied by the same log-normal factor, with median 1 and the coefficient of variation set by the level, so that its capacitance $\tau_\mathrm{mem}/R$ stays fixed; the factor applies to hidden and output neurons.
Beyond these four defects, we did not emulate drift over time, such as with temperature, or effects that span a whole array, such as voltage drops along its wires or the converters that read it.

\textbf{Hardware-in-the-loop training.} Each network trains and is tested at one level of one defect.
Weight noise is drawn anew for every batch, and $\tau$ mismatch once per network, as one simulated chip that stays fixed through training and test.
Under weight noise and quantization, the forward and backward passes use the perturbed or quantized weights, the update goes to the stored full-precision weights, and $F$ is recomputed from them at every update.
Validation also runs with the defect present, so it selects the epoch at which the network performs best with it; the noise drawn for validation leaves the training itself unchanged.

\textbf{Post-training perturbation.} We tested each method's ten defect-free networks with each defect added, without further training.

\textbf{Test chips.} Under weight noise in both settings, and under $\tau$ mismatch in post-training perturbation, each network is tested on 20 simulated chips, draws of the defect that it never met in training, shared by every network and method; Supplementary Methods gives the tests under the other defects.

\textbf{What each method measures.} Under membrane noise, weight noise and quantization, ChronoProp uses the FI curve of the defect-free neuron, measured once, without noise, before training.
Under $\tau$ mismatch in hardware-in-the-loop training, it measures each hidden neuron's own curve on its simulated chip before training, as in equation~\eqref{eq:fi} but neuron by neuron, and its backward pass simulates the output neurons with their actual time constants and resistances.
BPTT differentiates the exact noisy membrane potentials and every neuron's true $\tau_\mathrm{mem}$ and $R$, which a chip exposes at best approximately, so its results are an optimistic bound for BPTT trained through a chip.

\subsection*{Statistics}

Every condition of the accuracy comparisons was trained with ten seeds, a number we fixed before training.
Figs.~\ref{fig:temporal} and \ref{fig:adaptive} and Supplementary Fig.~\ref{sfig:dendritic} share one set of seeds, Figs.~\ref{fig:recipe} and \ref{fig:defects} another, and Fig.~\ref{fig:static} a third, so networks are paired by seed only within these groups.
No runs were excluded.
Values are mean $\pm$ s.d. over seeds, with $n-1$ in the denominator of the s.d.
We compared the methods with two-sided paired $t$-tests over seeds (over networks in Fig.~\ref{fig:defects}) and applied Holm's method~\cite{holm_1979} to each family of comparisons named in the legends.

\section*{Data availability}

MNIST~\cite{lecun_mnist}, Fashion-MNIST~\cite{xiao_fmnist}, SHD and SSC~\cite{cramer_shd} are publicly available.
The per-run results behind every figure, together with the defect-free networks of Fig.~\ref{fig:defects} and those trained at the defect levels quoted in the text, are available at Zenodo (\url{https://doi.org/10.5281/zenodo.24461579}).

\section*{Code availability}

The code is available at \url{https://github.com/Poirazi-Lab/chronoprop} under the MIT licence (\url{https://doi.org/10.5281/zenodo.24718667}), with documentation and tutorials that run in Google Colab at \url{https://poirazi-lab.github.io/chronoprop}.
It contains ChronoProp, the BPTT baseline, the scripts that train the SHD, MNIST and Fashion-MNIST networks and the scripts that recreate the figures from the archived results.

\section*{Acknowledgements}

We thank Spyridon Chavlis for comments on the manuscript.
This work was supported by the Stavros Niarchos Foundation (SNF) and the Hellenic Foundation for Research and Innovation (H.F.R.I.) under the 5th Call of ``Science and Society Action Always strive for excellence -- Theodoros Papazoglou'' (project no. 28056) and the European Union Horizon 2020 Research and Innovation Programme (101055340/ERC Cortical Coupling).

\par\vspace{\intextsep}
\begin{nolinenumbers}
\noindent\makebox[\textwidth][c]{\begin{minipage}{180mm}
{\setstretch{1.0}\small\noindent\textbf{Extended Data Table 1 \textbar\ Training methods for SNNs and how ChronoProp differs from each.}\par}
\vspace{4pt}
{\setstretch{1.0}\sffamily\fontsize{7}{8.6}\selectfont\setlength{\tabcolsep}{3pt}%
\noindent\begin{tabular}{@{}>{\raggedright\arraybackslash}p{30mm}>{\raggedright\arraybackslash}p{60mm}>{\raggedright\arraybackslash}p{85.5mm}@{}}
\toprule
\textbf{Method} & \textbf{How it works} & \textbf{How ChronoProp differs} \\
\midrule
BPTT with a surrogate gradient \cite{neftci} & Follows the gradient back through every simulation step, using a smooth function in place of the spike's derivative. It needs the membrane potential of every neuron at every step. & ChronoProp takes one gradient per time bin from spike counts, using the slope of a measured FI curve in place of the smooth function. \\
\arrayrulecolor{black!35}\specialrule{0.25pt}{2.5pt}{2.5pt}\arrayrulecolor{black}
Truncated BPTT \cite{guo_truncated} & Applies BPTT within separate time windows, each with its own loss and weight update. No gradient passes between windows. & ChronoProp also stops the gradient at each bin, but uses one slope per bin instead of one derivative per step, and combines all bins into one weight update. \\
\arrayrulecolor{black!35}\specialrule{0.25pt}{2.5pt}{2.5pt}\arrayrulecolor{black}
Rate-based backpropagation \cite{yu_ratebp} & Simulates step by step, keeping running averages of firing rates and of values based on a smooth derivative, then makes one backward pass through them. It is designed to approximate BPTT's gradient at lower cost. & Both pass the gradient through rates or counts instead of through every step. Rate-based backpropagation computes a smooth derivative from the membrane potential at every step, whereas ChronoProp reads spike counts and takes its slope from the FI curve measured on the neuron. It averages over the whole input and is designed to match BPTT's gradient, whereas ChronoProp takes one gradient per time bin and does not aim to match BPTT's. It trains feedforward networks on short input, whereas ChronoProp also trains recurrent networks on time-varying input. \\
\arrayrulecolor{black!35}\specialrule{0.25pt}{2.5pt}{2.5pt}\arrayrulecolor{black}
e-prop \cite{bellec_eprop} & Learns forward in time. Traces stored locally, built from spikes and the membrane potential at every step, are combined with an error signal sent to each neuron. & e-prop can learn online, as the input arrives. ChronoProp updates the weights once per batch, through a backward pass that needs only spike counts. \\
\arrayrulecolor{black!35}\specialrule{0.25pt}{2.5pt}{2.5pt}\arrayrulecolor{black}
EventProp \cite{wunderlich_eventprop} & Computes exact gradients by solving a second set of equations backward in time, with a jump at each spike. It needs spike times and the synaptic current at each spike, read or estimated. & ChronoProp's gradient is approximate and does not use spike timing within a bin, but it needs spike counts and a measured curve instead of the neuron's equations. \\
\arrayrulecolor{black!35}\specialrule{0.25pt}{2.5pt}{2.5pt}\arrayrulecolor{black}
DelGrad \cite{delgrad} & Computes exact gradients for weights and delays, using a formula derived from the neuron's equations for certain time constants. It records only spike times. & DelGrad learns from the timing of spikes. ChronoProp learns from spike counts, through a measured curve instead of a derived formula. \\
\arrayrulecolor{black!35}\specialrule{0.25pt}{2.5pt}{2.5pt}\arrayrulecolor{black}
Spike-train backpropagation \cite{zhang_strsbp} & Trains recurrent networks on whole spike trains. It computes the input each neuron receives from spike times, relates that input to the neuron's spike count, and solves a set of linear equations to pass the gradient through recurrent connections. & Both train recurrent networks without following the gradient through every step. Spike-train backpropagation uses the spike times of all neurons, whereas ChronoProp uses one count per neuron per time bin. Its relation comes from the firing threshold and a fit to random spike times, whereas ChronoProp's slope comes from the FI curve measured on the neuron. It solves linear equations for recurrence, whereas ChronoProp treats recurrent spikes as input counted in each bin. \\
\arrayrulecolor{black!35}\specialrule{0.25pt}{2.5pt}{2.5pt}\arrayrulecolor{black}
Tandem learning \cite{wu_tandem} & Feeds each spiking layer's spike counts, taken over the whole input, to an artificial layer with the same weights. The backward pass runs through the artificial layers, using a function derived from the neuron's equations. & Both use this pairing. Tandem learning derives its function from the neuron's equations, whereas ChronoProp fits it to the FI curve measured on the neuron. Tandem learning takes one count over the whole input, whereas ChronoProp takes one count per time bin. Tandem learning trains feedforward networks on short input, whereas ChronoProp also trains recurrent networks on time-varying input over many steps. \\
\arrayrulecolor{black!35}\specialrule{0.25pt}{2.5pt}{2.5pt}\arrayrulecolor{black}
Chip-in-the-loop training \cite{schmitt_hil} & Trains an artificial network first and copies its weights to the chip. Training then continues in a loop. The chip runs the forward pass, its firing rates are recorded, and software computes the gradient through a rectified linear function. & Both use a rectified linear function. This method assumes it, whereas ChronoProp fits its gain and threshold to the FI curve measured on the neuron. This method starts from weights trained on an artificial network, whereas ChronoProp trains from the start on spike counts. This method uses one firing rate per stimulus in feedforward networks on a small static task, whereas ChronoProp uses one count per time bin and also trains recurrent networks on time-varying input. \\
\arrayrulecolor{black!35}\specialrule{0.25pt}{2.5pt}{2.5pt}\arrayrulecolor{black}
Proxy learning \cite{kheradpisheh_proxy} & An artificial and a spiking network share their weights. The spiking network's output counts replace the artificial network's output, and the backward pass runs through the artificial network, using a rectified linear function assumed to match the neuron. & Both use a rectified linear function. Proxy learning assumes it, whereas ChronoProp fits its gain and threshold to the FI curve measured on the neuron. In proxy learning only the output counts come from the spiking network, so the hidden layers' values come from the artificial network, whereas ChronoProp uses the spike counts of every layer. Proxy learning counts over the whole input in feedforward networks, whereas ChronoProp counts per time bin and also trains recurrent networks. \\
\arrayrulecolor{black!35}\specialrule{0.25pt}{2.5pt}{2.5pt}\arrayrulecolor{black}
Neural Engineering Framework \cite{eliasmith_nef,braindrop} & Takes each neuron's response curve, computed from a model or measured, and solves the weights directly for a specified function. & ChronoProp also measures neurons, but trains hidden and recurrent weights on a task by gradient through the curve. \\
\bottomrule
\end{tabular}\par}
\end{minipage}}
\end{nolinenumbers}
\vspace{\intextsep}

\bibliographystyle{unsrtnat}
\bibliography{references}

\clearpage
\section*{Supplementary Information}

\subsection*{Supplementary Methods}

\subsubsection*{Operation count}

In the effective floating-point operations of Narduzzi et al.~\cite{narduzzi_eflop}, with the masks $M^{A}=[A\neq 0]$ and $M^{B}=[B\neq 0]$, a product of $A\in\mathbb{R}^{m\times k}$ and $B\in\mathbb{R}^{k\times n}$ and, for matrices of equal size, an element-wise product and a sum cost
\begin{align}
P_{AB} &= \sum_{i=1}^{m}\sum_{j=1}^{n}\max\Bigl(0,\;2\sum_{l=1}^{k}M^{A}_{il}M^{B}_{lj}-1\Bigr),
\label{eq:eflop_matmul}\\
P_{A\odot B} &= \sum_{i,j} M^{A}_{ij}\wedge M^{B}_{ij}, \qquad
P_{A+B} = \sum_{i,j} M^{A}_{ij}\vee M^{B}_{ij},
\label{eq:eflop_elem}
\end{align}
where a dot product with $q$ non-zero terms costs $q$ multiplications and $q-1$ additions.
We replayed each trained network's forward and backward passes operation by operation on its actual tensors, sample by sample, on the 10{,}000 MNIST test images and with the same input spikes for both methods, and applied these operators to the exact masks of the operands.
An activation function, a surrogate derivative or the mask of equation~\eqref{eq:proxy_grad} costs one operation per non-zero argument, a running sum one addition per non-zero term added, and the reset nothing; the output neurons, which never fire, are charged no threshold.
The replayed forward pass reproduces each network's output, and the replayed weight gradients match those of automatic differentiation to single-precision rounding.
Where a counting rule admits alternatives, such as charging a running sum for each added term or for its whole total, we computed the count under five conventions. The one we report, which charges each added term, also gives the smallest BPTT-to-ChronoProp ratio (4.17 at 2{,}000 steps, against 4.20--18.79 for the others).

\subsubsection*{Two-compartment neurons and the addSHD task}

The two-compartment neurons of Supplementary Fig.~\ref{sfig:dendritic} add a passive dendrite to the soma of equations~\eqref{eq:syn}--\eqref{eq:spike}, with its own synaptic current $I^\mathrm{d}_i$ and membrane potential $U^\mathrm{d}_i$ and the same time constants and resistance.
At every step, current flows between the two in proportion to the difference of their potentials,
\begin{align}
\hat U_i[t] &= \alpha\, U_i[t-1] + R\,(1-\alpha)\, I_i[t] - (1-\alpha)\bigl(U_i[t-1]-U^\mathrm{d}_i[t-1]\bigr), \nonumber\\
U^\mathrm{d}_i[t] &= \alpha\, U^\mathrm{d}_i[t-1] + R\,(1-\alpha)\, I^\mathrm{d}_i[t] + (1-\alpha)\bigl(U_i[t-1]-U^\mathrm{d}_i[t-1]\bigr),
\label{eq:dend}
\end{align}
and only the soma fires and resets.
To measure the FI plane, we counted the somatic spikes in one bin at constant currents into the soma and the dendrite, each at 101 values from 0 to 1.
The fit $f(I, I^\mathrm{d}) = \max\bigl(0,\; g\,(I-\theta_0) + g^\mathrm{d} I^\mathrm{d}\bigr)$ reaches $R^2 = 0.965$ at the fine timestep and 0.956 at the coarse one.
In the backward pass, each neuron has one input per compartment, built from the counts as in equation~\eqref{eq:proxy_in}, with the recurrent counts entering the soma's, and both are divided by $\kappa$ before the fitted plane is read.

In the addSHD task, each sample pairs an English digit from SHD, presented to the somata, with a German one, presented to the dendrites through separate input weights, and the label is the sum of the two digits (19 classes).
We held out 15\% of each language's training recordings for validation, as for SHD, and in training paired each English recording with a German one drawn anew every epoch, whereas the validation and test pairs are fixed and the same for every run.
The feedforward networks have 256 hidden neurons (363{,}264 parameters) and the recurrent ones 221, without self-connections (362{,}440 parameters).
Each method tried three learning rates around those selected in preliminary runs ($5\times10^{-5}$, $1.5\times10^{-4}$ and $5\times10^{-4}$ for ChronoProp; $3\times10^{-4}$, $10^{-3}$ and $3\times10^{-3}$ for BPTT).

\subsubsection*{Tests under hardware defects}

Under membrane noise, each network is tested with 5 noise draws, and under quantization once, since it is deterministic; under $\tau$ mismatch in hardware-in-the-loop training, it is tested once, on its simulated chip.
Under weight noise, we report only the tests on the 20 chips, because the test accuracy recorded during training overestimates the accuracy on new simulated chips; validation and test of one epoch share one noise draw, so the selected epoch tends to be one whose draw suited the network.

\clearpage
\subsection*{Supplementary Figures}

\begin{figure}[H]
  \boxsfig{figures/fig2_s1}{sfig:dtlimit}{\textbf{A coarse timestep distorts the neuron's responses and the FI curve that ChronoProp fits.} One current-based LIF (CUBA-LIF) neuron, simulated at $\Delta t = 0.1$~ms (\textbf{a},\textbf{c}; dark blue) and at $\Delta t = 1$~ms (\textbf{b},\textbf{d}; grey). In this figure, colour encodes the timestep.
\textbf{a},\textbf{b} Membrane potential in response to four identical input spikes 25~ms apart (triangles); the dotted line marks the firing threshold. At the fine timestep every response stays below threshold. At the coarse timestep, which is half the 2~ms synaptic time constant, the same input reaches the threshold and the neuron fires four times.
\textbf{c},\textbf{d} FI curve: spike count in a 50~ms window against a constant input current (dots; every second measured current is shown), with the fitted rectified linear function (ReLU; red line). Each panel prints the fit's $R^2$, gain and threshold, where the threshold is the current at which the line leaves zero. At the coarse timestep the count rises in wide, flat steps and saturates at a lower value, which lowers both the gain and $R^2$. The neuron starts firing at the same current for both timesteps, so the lower fitted threshold at the coarse timestep reflects the fit rather than the neuron.}
\end{figure}

\begin{figure}[H]
  \boxsfig{figures/fig2_s2}{sfig:staticdyn}{\textbf{ChronoProp and BPTT follow similar training trajectories.} Networks, conditions, markers and colours are as in Fig.~\ref{fig:static}b, and each method trains at its validation-selected learning rate.
\textbf{a},\textbf{b} Cross-entropy loss on the training (solid) and validation (dashed, lighter) sets for MNIST (\textbf{a}) and FMNIST (\textbf{b}), with one sub-panel per $\Delta t$ on a shared axis. Lines show means over $n = 10$ seeds and bands $\pm 1$~s.d. Panels \textbf{a} and \textbf{b} share the legend in \textbf{a}.
\textbf{c} Epoch of best validation accuracy, at which the reported model is taken, with one marker per seed; grey lines join the two methods' runs from the same seed (solid, MNIST; dashed, FMNIST). BPTT reaches its best epoch earlier at $\Delta t = 1$~ms ($P = 0.027$) but not at the finer timesteps (both $P \geq 0.67$; $n = 20$ seed pairs, datasets pooled; two-sided paired $t$-tests, Holm-corrected across the three timesteps).
\textbf{d} Test accuracy against learning rate (mean $\pm$ s.d. over $n = 10$ seeds; error bars are mostly smaller than the markers), with one sub-panel per $\Delta t$. Panels \textbf{c} and \textbf{d} share the legend in \textbf{d}.}
\end{figure}

\begin{figure}[H]
  \boxsfig{figures/fig2_s3}{sfig:activity}{\textbf{A closer look at ChronoProp's training cost and hidden-layer activity.} The figure extends Fig.~\ref{fig:static}c--f in two directions: \textbf{a} and \textbf{b} repeat the cost measurement with a spiking readout, and \textbf{c} and \textbf{d} follow the hidden-layer activity of Fig.~\ref{fig:static}f through training. Networks, step counts $T$ and colours are as in Fig.~\ref{fig:static}c--f, and the panels share the method legend in \textbf{a}.
\textbf{a} Training time per batch. The networks of Fig.~\ref{fig:static} were trained with a non-spiking readout and are only measured here with a spiking one.
\textbf{b} Peak GPU memory per batch (logarithmic axis).
\textbf{c} Mean hidden-layer firing rate during training at $T = 50$ (solid) and $T = 2{,}000$ (dashed), measured on the test images every 5 epochs from before the first update.
\textbf{d} Fraction of hidden neurons at epoch 60 that fire at least once in a test image (solid, mean over images) or never fire on any (dashed).
Green boxes (\textbf{a},\textbf{b}) give the BPTT-to-ChronoProp ratio at 1,000 and 2,000 steps. Error bars (\textbf{a},\textbf{b},\textbf{d}) and bands (\textbf{c}) show $\pm 1$~s.d. over $n = 3$ networks, each timed twice in \textbf{a}.}
\end{figure}

\begin{figure}[H]
  \boxsfig{figures/fig3_s4}{sfig:binning}{\textbf{A finer timestep resolves each bin's input in more detail and improves the fitted FI curve.} Dark blue marks $\Delta t = 0.1$~ms and grey $\Delta t = 1$~ms in \textbf{b},\textbf{c}.
\textbf{a} The same SHD recording at both timesteps, presented in 1~ms audio frames at $\Delta t = 0.1$~ms (left) and in 10~ms frames at $\Delta t = 1$~ms (right). In both cases the recording plays ten times faster than real time, so each $K = 10$~ms bin (dashed lines) covers the same 100~ms of audio (arrow) at both timesteps. The fine timestep presents each bin in 100 one-millisecond frames instead of 10 ten-millisecond ones, and so resolves the input in more detail.
\textbf{b} FI curve of the simulated neuron within one 10~ms bin at each timestep, measured at constant input currents (circles) and fitted with a rectified linear function (ReLU; red line).
\textbf{c} Fit quality ($R^2$) against bin duration.}
\end{figure}

\begin{figure}[H]
  \boxsfig{figures/fig3_s5}{sfig:temporaldyn}{\textbf{ChronoProp trains stably across network architectures and timesteps.} Training (solid) and validation (dashed) cross-entropy loss over epochs for temporal ChronoProp (tCP; blue) and BPTT (orange), in the feedforward (\textbf{a},\textbf{c}) and recurrent (\textbf{b},\textbf{d}) networks of Fig.~\ref{fig:temporal}c--f on SHD (\textbf{a},\textbf{b}) and SSC (\textbf{c},\textbf{d}). Each panel shows $\Delta t = 1$~ms on the left and 0.1~ms on the right. Lines show means over $n = 10$ seeds at each method's selected learning rate, and bands $\pm 1$~s.d. In \textbf{b}, the jump at epoch 229 ($\Delta t = 0.1$~ms) is one BPTT seed collapsing to chance; in \textbf{d}, the BPTT curves at $\Delta t = 0.1$~ms settle at $\ln 35$, the loss of a uniform prediction over the 35 classes. The panels share the legend in \textbf{a}.}
\end{figure}

\begin{figure}[H]
  \boxsfig{figures/fig4_s6}{sfig:adaptivedyn}{\textbf{With adaptive neurons, ChronoProp still trains stably, and BPTT's difficulties in recurrent networks at the fine timestep remain.} Training (solid) and validation (dashed) cross-entropy loss over epochs for temporal ChronoProp (tCP; blue) and BPTT (orange), in the adaptive feedforward (\textbf{a},\textbf{c}) and recurrent (\textbf{b},\textbf{d}) networks of Fig.~\ref{fig:adaptive} on SHD (\textbf{a},\textbf{b}) and SSC (\textbf{c},\textbf{d}). Each panel shows $\Delta t = 1$~ms on the left and 0.1~ms on the right. Lines show means over $n = 10$ seeds at each method's selected learning rate, and bands $\pm 1$~s.d. In \textbf{d}, the wide BPTT band at $\Delta t = 0.1$~ms spans seeds that collapse to $\ln 35$, the loss of a uniform prediction, and seeds that do not. The panels share the legend in \textbf{a}.}
\end{figure}

\begin{figure}[H]
  \boxsfig{figures/fig4_s7}{sfig:dendritic}{%
\textbf{ChronoProp extends to two-compartment neurons whose dendrite receives its own input.}
\textbf{a} In the addSHD task, each sample pairs two spoken digits from SHD, one in English and one in German. The English digit drives the soma of each hidden neuron and the German digit its passive dendrite, and the network must report their sum (here, $4 + 3 = 7$).
\textbf{b} FI plane of the two-compartment neuron: the spike count per bin, measured on the simulated neuron, against constant currents into the soma and the dendrite. The two compartments are coupled at every timestep, with current flowing in either direction in proportion to the difference between their membrane potentials. $R^2$ is the quality of fit of a rectified linear function of both currents, which ChronoProp differentiates in the backward pass.
\textbf{c},\textbf{d} Accuracy on addSHD, on the validation (\textbf{c}) and test (\textbf{d}) sets, for temporal ChronoProp (tCP; blue) and BPTT (orange) in feedforward (FF) and recurrent (REC) networks of these neurons with the same number of parameters, trained with $n = 10$ seeds and displayed as in Fig.~\ref{fig:temporal}c--f. The panels share the legend in \textbf{d}.
In two-sided paired $t$-tests over seeds, Holm-corrected across the four test comparisons, the two methods differ in test accuracy in every condition (all $P \le 1.9 \times 10^{-4}$).
}
\end{figure}

\begin{figure}[H]
  \boxsfig{figures/fig5_s8}{sfig:tricks}{\textbf{The three hardware-compatible techniques behind Fig.~\ref{fig:recipe}.}
\textbf{a} Receptive fields limit how many inputs each neuron receives. Input weights of 128 hidden neurons (columns) over 140 input channels (rows), the 700 SHD channels summed in groups of five. Each neuron connects to a contiguous window of 40 channels, spaced evenly across the range. Weights are illustrative random values (red, negative; blue, positive), and the black bar marks one neuron's window.
\textbf{b} Dropout silences a random set of hidden neurons in each training sample. Spikes of 128 hidden neurons over 80 simulation steps (synthetic data), with the kept neurons in dark blue and the dropped ones marked by pink bars. The spikes of a dropped neuron, still visible beneath its bar, are set to zero for the whole sample and reach neither the readout nor the recurrent connections.
\textbf{c} Channel jitter shifts each training sample along the channel axis by a random number of channels. One SHD training sample as recorded (left) and as the network receives it (right). The dark blue line marks the lower edge of the original's main activity and the pink line the same edge after the shift.}
\end{figure}

\begin{figure}[H]
  \boxsfig{figures/fig6_s9}{sfig:defectmech}{\textbf{The four hardware defects of Fig.~\ref{fig:defects}, illustrated.}
\textbf{a} Membrane noise with an s.d. of one firing threshold: the membrane potential of a neuron driven by a constant current to twice its threshold, over 50~ms, without and with noise. The dotted grey line marks the threshold.
\textbf{b} Weight noise at 10\% of full scale: programmed against target weight for a random subset of input synapses, with the identity line marking exact programming. Target weights beyond the full scale are clipped to it.
\textbf{c} $\tau$ mismatch with a coefficient of variation of 30\%: the distribution of membrane time constants across the hidden neurons of one simulated chip. The vertical line marks the design value of 10~ms. The spread arises in the leak conductance, so the membrane resistance scales with $\tau$ at fixed capacitance.
\textbf{d} Weight quantization at 4 bits: quantized against full-precision weight for all input synapses, which take 15 levels, with the identity line for reference. Weights beyond the outermost levels are clipped to them.
In all four panels, dark blue shows the ideal case and pink the non-ideal one, as in the legend in \textbf{b}.}
\end{figure}

\begin{figure}[H]
  \boxsfig{figures/fig6_s10}{sfig:defectptp}{\textbf{The hardware defects of Fig.~\ref{fig:defects}, imposed after defect-free training.}
\textbf{a} Membrane noise, in multiples of the firing threshold.
\textbf{b} Weight noise, in \% of full scale.
\textbf{c} $\tau$ mismatch, as the coefficient of variation of the membrane time constant.
\textbf{d} Weight quantization, with fewer bits towards the right.
The defect-free networks of Fig.~\ref{fig:defects} are tested with each defect added and no further training (post-training perturbation). Levels, axes, colours and the retained-accuracy scale match Fig.~\ref{fig:defects}, so the two figures can be compared point by point. Points show means and bands $\pm 1$~s.d. over networks and simulated chips ($n = 10$ networks per point, each scored on 20 chips in \textbf{b} and \textbf{c}, with 5 noise realizations in \textbf{a}, and once in \textbf{d}, where quantization is deterministic). The panels share the legend in \textbf{d}.}
\end{figure}

\end{document}